\documentclass{article}

\usepackage{microtype}
\usepackage{graphicx}
\usepackage{subcaption}
\usepackage{booktabs} %

\usepackage{hyperref}

\usepackage[preprint]{icml2026}

\usepackage{amsmath}
\usepackage{amssymb}
\usepackage{mathtools}
\usepackage{amsthm}

\usepackage[capitalize,noabbrev]{cleveref}

\theoremstyle{plain}

\theoremstyle{definition}

\theoremstyle{remark}

\usepackage[textsize=tiny]{todonotes}
\usepackage{multirow}

\begin{document}

\twocolumn[
  \icmltitle{STEP: Scientific Time-Series Encoder Pretraining via Cross-Domain Distillation}

  \icmlsetsymbol{equal}{*}
  \icmlsetsymbol{corresponding}{$\dagger$}
  \begin{icmlauthorlist}
    \icmlauthor{Chen Zhang}{equal,ailab,tsinghua}
    \icmlauthor{Liwei Liu}{equal,ailab,hit}
    \icmlauthor{Jun Tao}{equal,ailab,sichuan}
    \icmlauthor{Xiaoyu Yang}{cam}
    \icmlauthor{Xuenan Xu}{ailab}
    \icmlauthor{Kai Chen}{ailab}
    \icmlauthor{Bowen Zhou}{ailab,tsinghua}
    \icmlauthor{Wen Wu}{corresponding,ailab}
    \icmlauthor{Chao Zhang}{corresponding,ailab,tsinghua}
  \end{icmlauthorlist}

  \icmlaffiliation{ailab}{Shanghai Aritificial Intelligence Laboratory}
  \icmlaffiliation{tsinghua}{Tsinghua University}
  \icmlaffiliation{hit}{Harbin Institute of Technology}
  \icmlaffiliation{sichuan}{Sichuan University}
  \icmlaffiliation{cam}{University of Cambridge}

  \icmlcorrespondingauthor{Wen Wu}{wuwen@pjlab.org.cn}
  \icmlcorrespondingauthor{Chao Zhang}{cz277@tsinghua.edu.cn}

  \icmlkeywords{Machine Learning, ICML}

  \vskip 0.3in
]

\printAffiliationsAndNotice{}  %

\begin{abstract}
Scientific time series are central to scientific AI but are typically sparse, highly heterogeneous, and limited in scale, making unified representation learning particularly challenging. Meanwhile, foundation models pretrained on relevant time series domains such as audio, general time series, and brain signals contain rich knowledge, but their applicability to scientific signals remains underexplored.
In this paper, we investigate the transferability and complementarity of foundation models from relevant time series domains, and study how to effectively leverage them to build a unified encoder for scientific time series.
We first systematically evaluate relevant foundation models, showing the effectiveness of knowledge transfer to scientific tasks and their complementary strengths.
Based on this observation, we propose \textbf{STEP}, a \textbf{S}cientific \textbf{T}ime Series \textbf{E}ncoder \textbf{P}retraining framework via cross domain distillation.
STEP introduces adaptive patching to handle extreme-length sequences and a statistics compensation scheme to accommodate diverse numerical scales. It further leverages cross-domain distillation to integrate knowledge from multiple foundation models into a unified encoder. By combining complementary representations across different domains, STEP learns general-purpose and transferable features tailored for scientific signals.
Experiments on seven scientific time series tasks demonstrate that STEP provides both an effective structure and an effective pretraining paradigm, taking a STEP toward  scientific time series representation learning.
\end{abstract}

\section{Introduction}

Artificial intelligence for science (AI4Sci) has become one of the most active and impactful frontiers of modern AI research~\cite{alphafold, scidiscovery, bai2025intern}, aiming to accelerate scientific discovery by learning from complex observational and experimental data. Among various scientific data modalities, time series play a central role, arising in domains such as astronomy, geophysics, neuroscience, and bioacoustics. These signals encode rich temporal dynamics and physical processes, making them indispensable for understanding, modelling, and predicting scientific phenomena. Despite their importance, how to build unified and general-purpose representations for scientific time series remains largely underexplored.

Recent studies reveal that time series should be treated as a dedicated data modality~\cite{wu2025scitsscientifictimeseries}: serializing long temporal signals as text is inefficient and difficult to scale, while rasterizing them as images sacrifices numerical precision and struggles with high-dimensional signals.
Consequently, existing general-purpose multimodal LLMs, which typically rely on vision encoders and language models, are insufficient for capturing the richness of scientific time series. This motivates the design of specialized time series encoders.

Conventional time series encoders face additional challenges when applied to scientific data, primarily due to the extreme heterogeneity of scientific time series. Conventional time series encoders typically build on general time series such as sales, electricity load, or weather data~\cite{time_moe,woo2024moirai,zhou2021informer,NieNSK23}. These signals are typically sampled at coarse resolutions (\textit{e.g.}, minute-level or daily) and exhibit noticeable trends and seasonality. However, scientific temporal signals from different scientific domains differ drastically in their semantic content, sequence length, sampling frequency, numerical scale, number of channels, and noise characteristics. For example, scientific time series may range from short transient signals to ultra-long recordings, span sampling rates from micro-Hz to MHz, and vary from univariate measurements to high-dimensional multi-channel observations. 
In addition, foundation models for general time series are primarily trained to capture trends and seasonality for forecasting tasks. They focus on predicting overall temporal patterns rather than learning detailed representations of the underlying signal, limiting their performance on scientific tasks that require fine-grained, domain-specific information~\cite{wu2025scitsscientifictimeseries}. Furthermore, scientific data are typically sparse: while the number of signal categories is large, each category often has limited labelled samples, making large-scale pretraining particularly difficult.

Together, the extreme heterogeneity and data sparsity of scientific time series pose fundamental challenges for building encoders for  scientific time series. In this work, we take a STEP towards investigating these challenges by studying encoder architectures capable of handling highly diverse signals. In addition, given the limited availability of scientific data, we investigate whether foundation models pretrained on related time series domains, such as audio, general time series, and neural signals, can provide transferable knowledge to improve performance on scientific tasks.
We then propose \textbf{STEP}, a \textbf{s}cientific \textbf{t}ime-series \textbf{e}ncoder \textbf{p}retraining framework via cross-domain distillation. 
The STEP encoder introduces a novel adaptive patching mechanism to handle highly diverse sequence lengths and a statistics compensation scheme to accommodate varying numerical scales. It then leverages a cross-domain distillation framework to learn from multiple foundation models pretrained on different domains, effectively integrating their complementary knowledge.
Experiments on seven scientific time series tasks demonstrate that STEP establishes an effective pretraining paradigm, achieving strong performance across diverse scientific domains.

The main contributions are summarized as follows: (i) We systematically analyze the transferability of foundation models pretrained on related time series domains (\textit{i.e.}, audio, general time series, and neural signals), revealing the complementary strengths and limitations of different domains for scientific time series tasks.
(ii) We design the STEP encoder with adaptive patching and a statistics compensation scheme to handle highly heterogeneous scientific signals.
(iii) We adopt a cross-domain distillation framework for STEP that integrates knowledge from multiple foundation models, exploiting the complementary strengths of different domains.

\begin{figure*}[t]
    \centering
    \includegraphics[width=\linewidth]{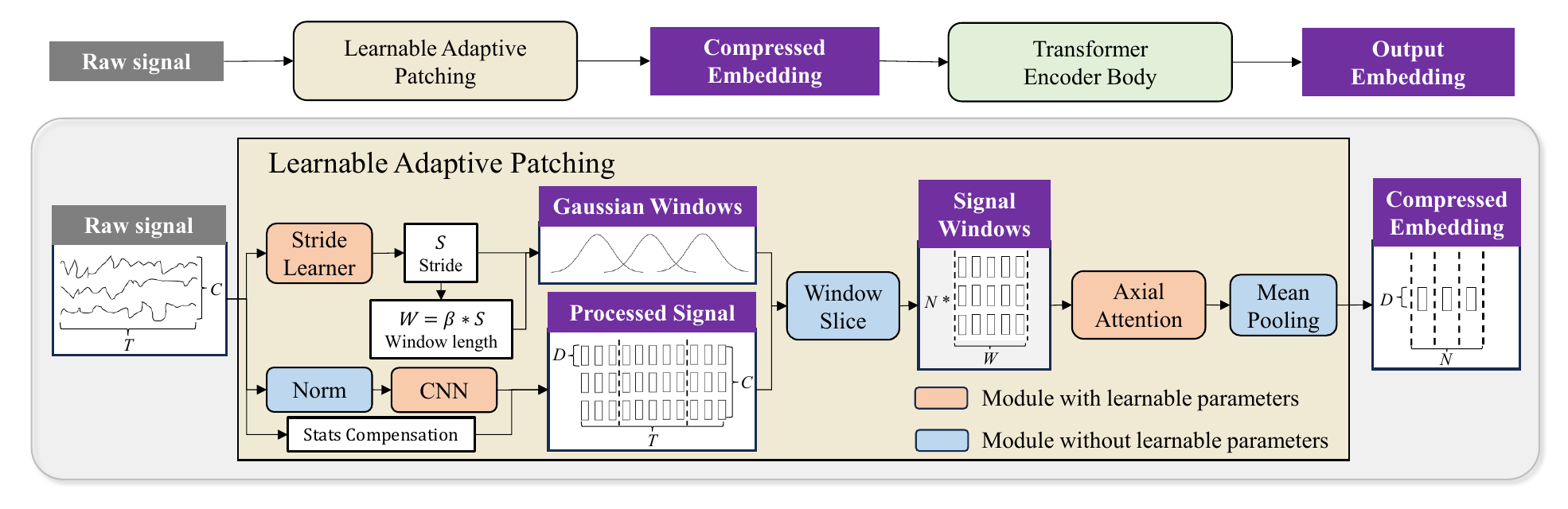}
    \caption{Architecture of the STEP Encoder. It includes an adaptive module that dynamically compresses the input signal and a statistics compensation scheme that recovers task-relevant statistical information.}
    \label{fig:subsample}
\end{figure*}

\section{Related work}

\subsection{Time Series Encoder Structures}

Recent advances in time series modelling commonly leverage Transformer-based architectures~\cite{vaswani2017attention}. But the quadratic complexity of self-attention poses significant challenges for long-sequence processing. To address this, works such as Informer~\citep{zhou2021informer}, Autoformer~\citep{wu2021autoformer}, FEDformer~\citep{pmlr-v162-zhou22g} introduces sparse attention to reduce computational cost to near-linear complexity. However, their relatively modest sequence compression and full-sequence point-wise attention still limit the scalability to extremely long input sequences.

A recent and effective approach, PatchTST~\citep{NieNSK23} and Moirai~\citep{woo2024moirai} segments the input into fixed-length temporal patches, reducing sequence length while enriching token semantics through local dynamics. TimeOmni~\citep{wu2025scitsscientifictimeseries} uses a router to select patch size and produce embeddings of appropriate length, allowing the handling of time series across diverse temporal scales. The reliance on fixed, manually tuned patch lengths still limits their adaptability as general encoders across varied sequence characteristics.

\subsection{Related Time Series Foundation Models}
This section summarises foundation models of relevant time series domains such as audio, general time series, and neural signals.

Audio is a broad and fundamental class of time series signals, and its foundation models have been extensively studied across large-scale acoustic data.
These models typically span speech~\cite{hsu2021hubertselfsupervisedspeechrepresentation, radford2023robust}, general audio (\textit{e.g.}, sound events)~\cite{huang2022amae,chen2022beatsaudiopretrainingacoustic}, and music~\cite{li2024mertacousticmusicunderstanding,zhu2025muqselfsupervisedmusicrepresentation}. 
Recent work has further explored multi-subdomain audio foundation models that unify speech and general audio within a single representation framework~\cite{yang2025spearunifiedsslframework}.
Acoustic signals are typically relatively high-frequency (\textit{e.g.}, 16kHz or 48kHz) and contain both periodic and aperiodic components. These models are usually pretrained on audio data at the order of hundreds of thousands of hours.

General time series represent a widely studied class of temporal signals, such as sales, electricity load, and weather data, typically sampled at coarse resolutions (\textit{e.g.}, daily) and characterized by strong trends and seasonality.
Foundation models for general time series are primarily focused on forecasting tasks, such as Moirai~\cite{woo2024moirai}, Chronos~\cite{ansari2024chronos}, TimesFM~\cite{das2024decoderonlyfoundationmodeltimeseries}, and TimeMoE~\cite{time_moe}. They are commonly pretrained on datasets containing tens to hundreds of billions of time points~\cite{woo2024moirai,ansari2024chronos,time_moe}. 

Another important class of time series is neural signals, such as electroencephalography (EEG) and magnetoencephalography (MEG), which are high-dimensional, multivariate recordings capturing brain activity at fine temporal and frequency resolutions, often with substantial noise.
Representative neural foundation models include LaBraM~\cite{jiang2024largebrainmodellearning} (pretrained on 2.5k hours EEG), CBraMod~\cite{wang2025cbramodcrisscrossbrainfoundation} (27k hours EEG), and BrainOmni~\cite{xiao2025brainomnibrainfoundationmodel} (2k hours EEG + 656 hours MEG). Those models focus on capturing spatio-temporal and frequency-specific representations.

Although pretrained models in these domains are well-established, scientific time series are far more diverse in length, frequency, and modality, motivating our study of whether pretrained models from these related domains can be effectively leveraged for scientific time series tasks.

\section{Methodology}

\subsection{Architecture of the STEP Encoder}

To handle extremely diverse sequence lengths and heterogeneous channel dimensions, we design an encoder that adaptively compresses inputs into compact representations without handcrafted heuristics. The compression ratio is learned end-to-end, allowing the model to automatically determine an appropriate downsampling strategy for each sequence, while projecting inputs with different channel counts into a unified embedding space.

Our framework adopts a two-stage training pipeline and centres on a Learnable Adaptive Patching module for efficient long-sequence modelling. The architecture consists of three components: a learnable adaptive patching block, a Transformer encoder backbone, and
projection layers for distillation and downstream tasks. The overall structure is illustrated in Figure~\ref{fig:subsample}.

\textbf{Learnable Adaptive Patching. } 
To enable adaptive handling of highly diverse sequence lengths and channel dimensions, we introduce a learnable adaptive patching block that replaces fixed patching with a differentiable, learnable windowing mechanism.

Given an input time series $ \mathbf{X} \in \mathbb{R}^{T \times C}$, where $T$ denotes sequence length and $C$ denotes number of channels, the module first applies a channel-independent 1D convolution to obtain $ \mathbf{X'} \in \mathbb{R}^{T \times C \times D}$, where $D$ denotes hidden dimension. A set of Gaussian windows $ \mathbf{G} \in \mathbb{R}^{N \times W }$ is then used to segment $ \mathbf{X'} $ into overlapped patches $ \mathbf{X_\text{patch}} \in \mathbb{R}^{N \times W \times C \times D }$ via differentiable weighted aggregation, where $N$ is the number of windows parameterized by learnable stride $S$ and window length $W$. Axial attention~\cite{Ho2019AxialAI} is applied sequentially along the temporal and channel dimensions within each patch to capture local spatio-temporal structure, followed by mean pooling to yield a downsampled embedding $ \mathbf{E} \in \mathbb{R}^{N \times D }$.

Specifically, the stride learner direct maps the input sequence to the stride parameter via a multilayer perceptron (MLP): 
$
S = 1 + \exp\!\left( \operatorname{MLP}\left( \log(T) \right) \right)
$,
with logarithmic transformation applied to ensure numerical stability, flexible adaptation to diverse signal durations, and strictly positive strides for valid temporal downsampling. To prevent unbounded window length growth, motivated by the model’s incentive to maximize per-step information, we fix the window length to twice the stride (\textit{i.e.}, $ \beta=2 $). To further ensure the stability of striding learning, we adopt a stride penalty and a length penalty defined as follows:
\begin{align*}
    \text{stride penalty} &= \operatorname{huber}\!\left(\max\left(0,\, \log(S) - \log(S_{max})\right)\right), \\
    \text{length penalty} &= \operatorname{huber}\!\left(\max\left(0,\, \log(N') - \log(N_{max})\right)\right) \\
    &\quad + \operatorname{huber}\!\left(\max\left(0,\, \log(N_{min}) - \log(N')\right)\right).
\end{align*}
where $N'$ is a differentiable estimate of the embedding length $N$. In our experiments, $S_{max}$ is set as 400, $N_{max}$ is set as 200, $N_{min}$ is set as 5.

\textbf{Transformer Encoder Body}: The architecture of the Transformer encoder body follows the design of the Whisper encoder~\cite{radford2023robust}. Specifically, the input time series subsampled features are first processed by two 1D convolutional layers. The resulting representations are then augmented with learnable positional embeddings and fed into 17 Transformer encoder blocks of hidden dimension 768.

\textbf{Statistics Compensation Scheme}: The numerical scales of the input model often vary by orders of magnitude. Global normalization would induce unstable training dynamics, whereas dataset-specific normalization requires prior knowledge of data source at inference time. Moreover, statistics such as mean and variance can encode useful task-relevant information.
Therefore, we adopt a per-sample, per-channel normalization strategy: each input channel is independently standardized to zero mean and unit variance. The computed channel-wise statistics are log-scaled  and concatenated as auxiliary features to the output of the initial convolutional layers, as shown in Figure~\ref{fig:subsample}. This design preserves discriminative signal-level statistics while ensuring consistent input scale across heterogenous signals.

\subsection{Cross-Domain Knowledge Distillation}

\begin{figure}[t]
    \centering
    \includegraphics[width=\linewidth]{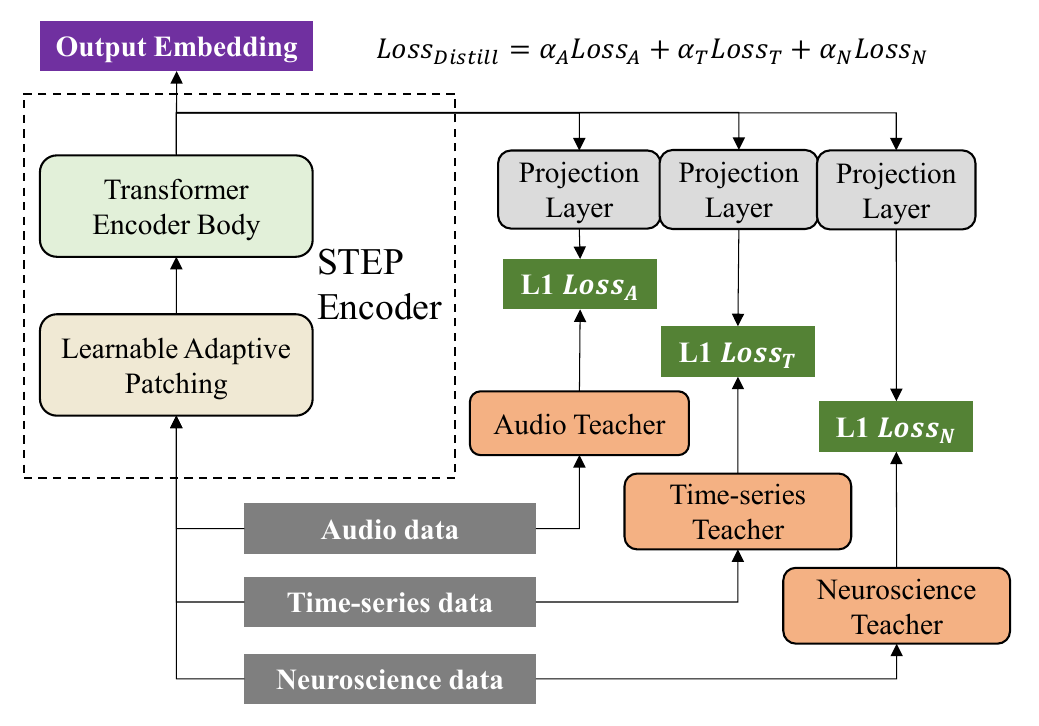}
    \caption{Illustration of the multi-teacher distillation process. }
    \label{fig:distill}
    \vspace{-1ex}
\end{figure}

To effectively leverage the rich representations learned by  pretrained models from related domains, we propose a cross-domain knowledge distillation framework. As illustrated in Figure~\ref{fig:distill}, the STEP encoder is pretrained by distilling from multiple teachers cross different relevant domain. 
Pretraining data is fed to both the corresponding teacher and the STEP encoder. The teacher’s output features serve as supervision targets, with projection layers aligning the encoder’s outputs to the teacher’s feature space.
The distillation losses from all teachers are then combined.
During the distillation phase, the stride is forced such that the output sequence length of the student encoder aligns with the teacher model.

 \begin{table*}[t]
\centering
\caption{Overview of downstream scientific datasets and corresponding tasks. ``Len.'' denotes length,``Freq.'' denotes frequency,  ``Ch.'' denotes channels.}
\resizebox{\textwidth}{!}{%
\small
\begin{tabular}{lcccccc}
\toprule
\textbf{Dataset} & \textbf{Discipline} & \textbf{\# Sample} & \textbf{Len. Range} & \textbf{Freq. (Hz)} & \textbf{\# Ch.} & \textbf{Task} \\
\midrule
GWOSC~\citep{wu2025scitsscientifictimeseries} & Astronomy & 1024 & 1.6e4 & 6.4e4 & 1 & Gravitational wave detection \\
LEAVES~\citep{Fei_2024} & Astronomy &1024 & 24$\sim$1.3e3 & 1.2e-5 & 1 & Light curve recognition \\
STEAD~\citep{mousavi2019stanford} & Earth Science & 2048 & 1.0e2 &6.0e3 & 3 & Earthquake detection  \\
MarmAudio~\citep{lamothe_2025_15017207} & Bioacoustics &900 & 6.2e4$\sim$2.4e5 & 9.6e4 & 1 & Vocalisation recognition \\
SleepEDF~\citep{Kemp2000SleepEDF} & Neuroscience & 1024 & 3.0e3 & 1.0e2 & 1 & Sleep staging \\
WBCIC~\citep{Yang_Rong_2024}  & Neuroscience &3600 & 1.0e3 & 2.5e2 & 58 & Motor imagery classification \\
RadSeg~\citep{huang2024radseg} & Radar &963 & 33$\sim$3.2e2 & 3.2e6 & 2  & Coding scheme recognition \\
\bottomrule
\end{tabular}%
}  %
\label{tab:downstream_datasets}
\end{table*}

\section{Experimental Setup}

\subsection{Downstream Scientific Tasks and Datasets}
\label{sec: downstream intro}
The proposed method is evaluated on seven downstream scientific tasks spanning astronomy, earth science, bioacoustics, neuroscience, and radar. These tasks cover a broad range of signal modalities and temporal dynamics, as summarized in Table~\ref{tab:downstream_datasets}. The tasks span sequence lengths from $10^1$ to $10^5$ and frequency ranges from $10^{-6}$ to $10^6$, covering both univariate and multivariate signals. 

The \textbf{GWOSC}~\cite{wu2025scitsscientifictimeseries} dataset is a synthetic gravitational-wave dataset generated following established simulation-and-injection pipelines in astronomy studies~\cite{dax2021real,dax2023neural,dax2025real}, where numerically simulated gravitational-wave signals are injected into real strain data from the GW Open Science Center (GWOSC). The downstream task involves identifying whether merge event happens in the gravitational wave.

\textbf{LEAVES}~\cite{Fei_2024} is a light-curve dataset developed for automatic variable star classification. It integrates publicly available observations from three major astronomical surveys— the All-Sky Automated Survey for SuperNovae (ASAS-SN), Gaia, and the Zwicky Transient Facility (ZTF). The downstream task involves categorizing variable stars into seven superclasses based on the observed light curve: Eclipsing binaries, Rotational variables, RR Lyrae, Cepheids, Long-period variables, Delta Scuti, and Non-variable.

\textbf{STEAD}~\cite{mousavi2019stanford} is a global dataset containing earthquakes that occurred between January 1984 and August 2018. All waveforms from the dataset have three channels, each of 1-minute length with a sampling rate of 100Hz. The task is to identifying whether  earthquake happens from seismic data. 

\textbf{MarmAudio}~\cite{lamothe_2025_15017207} is a dataset of common marmoset vocalizations, recorded over 40 months at a sampling rate of 96 kHz in a soundproofed animal facility room containing three cages. The dataset is annotated with six vocalization types: Infantcry, Phee, Seep, Trill, Tsik, and Twitter.

\textbf{WBCIC-SHU}~\cite{Yang_Rong_2024} dataset is a comprehensive motor imagery brain computer interface dataset. The tasking involves decoding the brain signal into three classes: left-hand movement, right-hand movement, foot movement.

\textbf{SleepEDF}~\cite{Kemp2000SleepEDF} is a comprehensive dataset containing overnight polysomnography (PSG) recordings. Each recording includes signals of two-channel EEG (Fpz-Cz and Pz-Cz) and one EOG, sampled at 100Hz. We follow prior work and process the data by utilizing only the Fpz-Cz EEG channel and perform 5-class sleep staging: W, N1, N2, N3, and REM. 
The EEG signals are processed with a 0.5-45Hz bandpass Butterworth filter to remove background noise.

\textbf{Radar Segmentation Dataset (RadSeg)}~\cite{huang2024radseg} is a synthetic radar dataset designed for building semantic segmentation models for radar activity recognition. RadSeg provides sample-wise annotations for interleaved radar pulse activities that extend across a long time horizon. RadSeg contains pulsed radar signals at varying signal-to-noise ratios between -20 and 20 dB with a resolution of 0.5 dB. The downstream task is to identify ground targets or terrain types.

\subsection{Baselines and Foundation Models Investigated}
\label{sec: baselines}
The proposed STEP encoder structure was compared to the following SOTA models: {PatchTST}~\cite{NieNSK23}, {Informer}~\cite{zhou2021informer}, Moirai~\cite{woo2024moirai}, TimeMoE~\cite{time_moe}. We investigate the transferability of the following foundation models to scientific time series:

\textbf{Audio foundation models}: (i) Whisper\footnote{\url{https://huggingface.co/openai/whisper-large-v3}}~\cite{radford2023robust} pretrained on 680k hours of multilingual labelled speech data through weak supervision; (ii) SPEAR\footnote{\url{https://huggingface.co/marcoyang/spear-xlarge-speech-audio}}~\cite{yang2025spearunifiedsslframework} pretrained on a heterogeneous mixture of speech and general audio data.

\textbf{General time series foundation models}: (i) TimeMoE\footnote{\url{https://huggingface.co/Maple728/Time-MoE-200M}}~\cite{time_moe}, a decoder-based time series forecasting model pretrained on over 300 billion time points spans over 9 domains; (ii) Moirai\footnote{\url{https://huggingface.co/Salesforce/moirai-1.1-R-large}}~\cite{woo2024moirai}, an encoder-based time series forecasting model pretrained on 27B data spanning 9 domains. 

\textbf{Neural foundation models}: (i) CBraMod\footnote{\url{https://huggingface.co/weighting666/CBraMod}}~\cite{wang2025cbramodcrisscrossbrainfoundation} an EEG foundation model pretrained on the Temple University Hospital EEG Corpus (TUEG)~\cite{tueg9353647}; (ii) BrainOmni\footnote{\url{https://huggingface.co/OpenTSLab/BrainOmni}}~\cite{xiao2025brainomnibrainfoundationmodel} which unifies EEG and MEG signals,  pretrained on 2k hours of EEG and 656 hours of MEG data.

\subsection{Implementation Details}
Pretrained models were finetuned for scientific tasks by adding three layers of MLP after their hidden layers. Multi-channel data is flattened into one dimension for models that do not support multi-channel input. Training set of each downstream task contains 20k samples and 4k samples were reserved for validation. The batch size is set to 32, and weighted cross-entropy loss is employed to address class imbalance. 
The procedure for full finetuning STEP is described as follows. During the first 1000 steps, only the Learnable Adaptive Patching module is optimized while the Transformer encoder body and classification head are frozen, serving as a warm-up for learning stable adaptive subsampling. From 1000 to 2000 steps, the adaptive patching module and the downstream MLP head are jointly trained with the encoder kept frozen. After 2000 steps, all components are fully unfrozen and optimized end-to-end.
Audio pretraining data uses LibriSpeech 960h dataset~ \citep{7178964}. General time series pretraining uses
Time-300B~\citep{time_moe}. Neural signal pretraining follows the setup of  BrainOmni~\cite{xiao2025brainomnibrainfoundationmodel}.

\begin{table*}[htbp]
  \centering
  \caption{The Accuracy (\%) and F1 score (\%) of foundation models from relevant domains on downstream scientific time series tasks. ``{PT}'' refers to whether pretrain weights are loaded. ``{FT}'' refers to whether encoder is finetuned during training. Best results in each column shown in bold. Second best underscored.}
  \label{tab:Transferability_performance_time_series}
  \resizebox{\textwidth}{!}{
  \begin{tabular}{c c c c c c c c c c c c c c c c c c} %
    \toprule
    \midrule
    \multirow{2}{*}{\textbf{Model}} 
    & \multirow{2}{*}{\textbf{PT}} 
    & \multirow{2}{*}{\textbf{FT}} 
    & \multicolumn{2}{c}{\textbf{GWOSC}} 
    & \multicolumn{2}{c}{\textbf{LEAVES}} 
    & \multicolumn{2}{c}{\textbf{STEAD}}
    & \multicolumn{2}{c}{\textbf{MarmAudio}} 
    & \multicolumn{2}{c}{\textbf{SleepEDF}} 
    & \multicolumn{2}{c}{\textbf{WBCIC}}
    & \multicolumn{2}{c}{\textbf{RadSeg}}  \\
    \cmidrule(lr){4-5} 
    \cmidrule(lr){6-7} 
    \cmidrule(lr){8-9} 
    \cmidrule(lr){10-11} 
    \cmidrule(lr){12-13} 
    \cmidrule(lr){14-15} 
    \cmidrule(lr){16-17}
    
    & &
    & \textbf{Acc} {$\uparrow$} & \textbf{F1} {$\uparrow$} 
    & \textbf{Acc} {$\uparrow$} & \textbf{F1} {$\uparrow$} 
    & \textbf{Acc} {$\uparrow$} & \textbf{F1} {$\uparrow$} 
    & \textbf{Acc} {$\uparrow$} & \textbf{F1} {$\uparrow$} 
    & \textbf{Acc} {$\uparrow$} & \textbf{F1} {$\uparrow$} 
    & \textbf{Acc} {$\uparrow$} & \textbf{F1} {$\uparrow$} 
    & \textbf{Acc} {$\uparrow$} & \textbf{F1} {$\uparrow$} \\
    
    \midrule
    \multirow{3}{*}{Whisper} 
    & $\surd$ & $\times$
    & 58.5 & 53.2 
    & 65.2 & 33.2 
    & 97.9 & 97.9  
    & 70.7 & 65.7 
    & 64.8 & 46.8 
    & 34.5 & 27.6
    & 43.0 & 37.4 \\
    
    & $\surd$ & $\surd$ 
    & 97.9 & 97.9 
    & 81.9 & 48.7 
    & \textbf{99.9} & \textbf{99.9} 
    & \textbf{97.9} & \textbf{97.9} 
    & \textbf{80.5} & \underline{71.1} 
    & 33.3 & 16.7
    & 78.4 & 78.6\\
    
    & $\times$ & $\surd$ 
    & 97.5 & 97.5 
    & 82.0 & 51.4 
    & 99.6 & 99.6 
    & 96.7 & 96.7 
    & \underline{77.8} & 64.9 
    & 33.3 & 16.7
    & 69.7 & 69.4 \\

    \midrule
    \multirow{3}{*}{SPEAR} 
    & $\surd$ & $\times$
    & 88.6 & 88.4 
    & 62.2 & 43.2 
    & 93.1 & 93.1 
    & 82.9 & 81.7 
    & 70.9 & 60.2 
    & 36.4 & 29.5
    & 58.4 & 58.3 \\
    
    & $\surd$ & $\surd$  
    & \textbf{99.5} & \textbf{99.5} 
    & 71.6 & 56.3 
    & \underline{99.8} & \underline{99.8} 
    & \underline{97.6} & \underline{97.5} 
    & 73.5 & 64.6 
    & 40.8 & 35.0
    & 78.4 & 78.4 \\
    
    & $\times$ & $\surd$ 
    & 96.6 & 96.6 
    & 74.3 & 65.9 
    & 99.7 & 99.7 
    & 94.0 & 93.9 
    & 76.7 & 70.2 
    & 34.8 & 23.6
    & 82.8 & 82.8 \\

    \midrule
    \midrule
    \multirow{3}{*}{TimeMoE} 
    & $\surd$ & $\times$
    & 66.9 & 63.9 
    & 83.8 & \underline{81.2} 
    & 84.1 & 83.7 
    & 52.1 & 52.4 
    & 75.5 & 65.4 
    & 34.1 & 20.8
    & 85.6 & 85.5 \\
    
    & $\surd$ & $\surd$  
    & 97.3 & 97.3 
    & \textbf{89.3} & \textbf{83.9} 
    & 97.9 & 97.9 
    & 57.9 & 59.0 
    & \textbf{80.5} & \textbf{72.2} 
    & 36.7 & 27.0
    & \underline{93.8} & \underline{93.8} \\
    
    & $\times$ & $\surd$ 
    & 78.7 & 77.8 
    & 82.0 & 72.7 
    & 92.8 & 92.7 
    & 50.7 & 50.6 
    & 69.3 & 57.1 
    & 33.9 & 21.2
    & 86.5 & 86.5 \\

    \midrule
    \multirow{3}{*}{Moirai} 
    & $\surd$ & $\times$ 
    & 58.1 & 56.6 
    & 73.8 & 62.3 
    & 96.0 & 96.0 
    & 46.0 & 44.4 
    & 64.4 & 54.5 
    & 33.9 & 27.0
    & 70.5 & 70.0 \\
    
    & $\surd$ & $\surd$  
    & \underline{98.2} & \underline{98.2} 
    & \underline{87.5} & 80.6 
    & 99.7 & 99.7 
    & 95.4 & 95.5
    & 77.4 & 70.4 
    & 32.4 & 25.8
    & \textbf{94.4} & \textbf{94.4} \\
    
    & $\times$ & $\surd$ 
    & 80.0 & 80.0 
    & 79.3 & 57.8 
    & 99.6 & 99.6 
    & 85.1 & 85.2 
    & 73.7 & 66.2 
    & 33.3 & 16.7
    & 87.6 & 87.8 \\

    \midrule
    \midrule
    \multirow{3}{*}{CBraMod} 
    & $\surd$ & $\times$
    & 53.1 & 51.3 
    & 51.5 & 13.1 
    & 88.6 & 88.6 
    & 26.0 & 16.3 
    & 72.8 & 63.5 
    & 49.9 & 40.1
    & 48.9 & 44.9 \\
    
    & $\surd$ & $\surd$  
    & 81.8 & 81.7 
    & 76.4 & 65.3 
    & 99.3 & 99.3 
    & 72.8 & 72.1 
    & 77.0 & 69.2 
    & \textbf{60.5} & \textbf{52.6}
    & 88.3 & 88.4 \\
    
    & $\times$ & $\surd$
    & 56.4 & 54.9 
    & 60.5 & 44.4 
    & 95.9 & 95.9 
    & 33.7 & 25.7 
    & 75.1 & 68.0 
    & \underline{57.2} & \underline{49.0}
    & 80.1 & 80.3 \\
    \midrule
    \bottomrule
  \end{tabular}
  }
\end{table*}

\section{Transferability of Pretrained Models to Scientific Tasks}
\label{sec: transfer}

We first examine how pretrained foundation models of related domain (see Section~\ref{sec: baselines}) perform on scientific time series tasks (introduced in Section~\ref{sec: downstream intro}).
Specifically, we test three training setups:
(i) Loading the pretrained weights but keep the encoder fixed, training only the downstream block; (ii) Loading the pretrained weights and finetune both the encoder and downstream block; (iii) All weights are randomly initialized and trained from scratch, serving as a baseline to evaluate the effectiveness of the backbone architecture.
Results are shown in Table~\ref{tab:Transferability_performance_time_series}.

\textbf{Pretrained encoder frozen \textit{vs.} finetuned.} First, comparing the first two rows for each model, we observe that finetuning the encoder generally improves performance over keeping it frozen, as expected.
But for high-dimensional scientific temporal tasks (\textit{i.e.}, WBCIC), Whisper and Moirai show opposite trends possibly because these models can only accept one-dimensional data input. When high-dimensional data is flattened into one dimension, the channel related information inside is destroyed, and the full finetuning is more prone to instability.
Keeping the pretrained weights frozen leads to near-chance performance on several tasks, indicating that existing pretrained models do not readily support direct transfer to heterogeneous scientific domains.

\textbf{Pretrained weights \textit{vs.} random initialization.} For the strategies that finetune the pretrained encoder versus training from scratch (comparing last two rows for each model), the performance difference reflects the effectiveness of the pretrained weights. For example, for SPEAR, finetuning the pretrained encoder improves performance on GWOSC, MarmAudio, and WBCIC tasks, indicating that the pretrained weights are beneficial for these tasks. In contrast, performance on LEAVES, SleepEDF, and RadSeg tasks is worse than training from scratch, while results on STEAD are comparable, suggesting that the pretrained knowledge of SPEAR transfers poorly to these tasks. These results indicate that the effectiveness of pretrained weights varies across different scientific tasks.

\textbf{Intra-domain analysis.} Among audio foundation models, SPEAR and Whisper excel on different tasks.
Comparing general time series models, it is observed that TimeMoE gives much worse results on MarmAudio dataset compared to Moirai. It is possibly due to the input length restriction. Moirai divides the input into multiple patches for computation, allowing it to handle longer time series data while TimeMoE has a maximum input length of 4096.

\textbf{Cross-domain comparison.} 
Overall, pretrained models from different domains exhibit varying effectiveness on scientific time series tasks, reflecting task-specific biases. Whisper and SPEAR perform best on acoustic-related tasks such as marmoset vocalisation classification (MarmAudio). For TimeMoE and Moirai, perform best on relatively short sequences such as LEAVES and RadSeg, but struggling with high-dimensional data such as WBCIC. CBraMod demonstrates superior performance on the WBCIC task, which can be attributed to its extensive pretraining on the TUEG dataset and the incorporation of mechanism that effectively integrates spatial and temporal representations.
BrainOmni adopts a two-stage training pipeline consisting of a neural tokenizer and a foundation model, which prevents straightforward end-to-end finetuning across domains and is therefore excluded from the table.

\begin{table*}[t]
  \centering
  \caption{Baseline comparison of encoder architectures. All models trained from scratch. Best results in each column shown in bold. Second best underscored. All values  reported in percentage (\%).}
  \label{tab:baseline struc}
  \resizebox{0.9\textwidth}{!}{
  \begin{tabular}{c c c c c c c c c c c c c c c c}
    \toprule
    \multirow{2}{*}{\textbf{Model}} 
    & \multicolumn{2}{c}{\textbf{GWOSC}} 
    & \multicolumn{2}{c}{\textbf{LEAVES}} 
    & \multicolumn{2}{c}{\textbf{STEAD}} 
    & \multicolumn{2}{c}{\textbf{MarmAudio}} 
    & \multicolumn{2}{c}{\textbf{SleepEDF}} 
    & \multicolumn{2}{c}{\textbf{WBCIC}} 
    & \multicolumn{2}{c}{\textbf{RagSeg}} \\
    
    \cmidrule(lr){2-3} 
    \cmidrule(lr){4-5} 
    \cmidrule(lr){6-7} 
    \cmidrule(lr){8-9} 
    \cmidrule(lr){10-11} 
    \cmidrule(lr){12-13} 
    \cmidrule(lr){14-15}
    
    & \textbf{Acc} {$\uparrow$} & \textbf{F1} {$\uparrow$} 
    & \textbf{Acc} {$\uparrow$} & \textbf{F1} {$\uparrow$}  
    & \textbf{Acc} {$\uparrow$} & \textbf{F1} {$\uparrow$}  
    & \textbf{Acc} {$\uparrow$} & \textbf{F1} {$\uparrow$}  
    & \textbf{Acc} {$\uparrow$} & \textbf{F1} {$\uparrow$}  
    & \textbf{Acc} {$\uparrow$} & \textbf{F1} {$\uparrow$}  
    & \textbf{Acc} {$\uparrow$} & \textbf{F1} {$\uparrow$}  \\
    \midrule

    PatchTST 
    & 74.8 & 72.5 
    & \underline{86.1} & 72.9 
    & 83.7 & 83.3 
    & 67.2 & 62.6 
    & \textbf{79.5} & \underline{67.7}
    & {47.3} & {38.9} 
    & 87.6 & 87.7 \\

    Informer 
    & 58.7 & 50.3  
    & 75.0 & \underline{74.7}
    & 74.5 & 74.7
    & 51.8 & 45.3
    & 74.0 & 61.5  
    & \textbf{54.6} & \textbf{43.7} 
    & \underline{92.5} & \underline{92.5}\\

    TimeMoE
    & 78.7 & 77.8 
    & 82.0 & 72.7 
    & 92.8 & 92.7 
    & 50.7 & 50.6 
    & 69.3 & 57.1 
    & 33.9 & 21.2
    & 86.5 & 86.5 
    \\
    Moirai 
    & \underline{80.0} & \underline{80.0} 
    & 79.3 & 57.8 
    & \textbf{99.6} & \textbf{99.6} 
    & \underline{85.1} & \underline{85.2} 
    & 73.7 & 66.2 
    & 33.3 & 16.7
    & 87.6 & 87.8 
    \\

     STEP 
    & \textbf{97.3} & \textbf{97.4}
    & \textbf{92.2} & \textbf{91.2}
    & \underline{99.0} & \underline{99.0} 
    & \textbf{94.7} & \textbf{94.6} 
    & \underline{78.2} & \textbf{76.3} 
    & \underline{48.7} & \underline{43.2} 
    & \textbf{95.2} & \textbf{95.2} \\
    
    \bottomrule
  \end{tabular}
  }
\end{table*}

\begin{table*}[t]
  \centering
  \caption{Ablation results of STEP, trained from scratch. Results with distinct decrease are marked by asterisk. All values  reported in percentage (\%).}
  \setlength{\tabcolsep}{5pt}
  \resizebox{\textwidth}{!}{
  \begin{tabular}{l c c c c c c c c c c c c c c} 
    \toprule
      & \multicolumn{2}{c}{\textbf{GWOSC}} 
      & \multicolumn{2}{c}{\textbf{LEAVES}} 
      & \multicolumn{2}{c}{\textbf{STEAD}} 
      & \multicolumn{2}{c}{\textbf{MarmAudio}} 
      & \multicolumn{2}{c}{\textbf{SleepEDF}} 
      & \multicolumn{2}{c}{\textbf{WBCIC}} 
      & \multicolumn{2}{c}{\textbf{RadSeg}} \\
      
    \cmidrule(lr){2-3} 
    \cmidrule(lr){4-5} 
    \cmidrule(lr){6-7} 
    \cmidrule(lr){8-9} 
    \cmidrule(lr){10-11} 
    \cmidrule(lr){12-13} 
    \cmidrule(lr){14-15}
    
     \textbf{STEP} & \textbf{Acc} {$\uparrow$} & \textbf{F1} {$\uparrow$} 
     & \textbf{Acc} {$\uparrow$} & \textbf{F1} {$\uparrow$} 
     & \textbf{Acc} {$\uparrow$} & \textbf{F1} {$\uparrow$} 
     & \textbf{Acc} {$\uparrow$} & \textbf{F1} {$\uparrow$} 
     & \textbf{Acc} {$\uparrow$} & \textbf{F1} {$\uparrow$} 
     & \textbf{Acc} {$\uparrow$} & \textbf{F1} {$\uparrow$} 
     & \textbf{Acc} {$\uparrow$} & \textbf{F1} {$\uparrow$} \\
    \midrule
    $-$ Adaptive patching
    & \textit{73.1}$^*$ & \textit{73.2}$^*$
    & 92.3 & 90.0 
    & 98.9 & 98.9 
    & 93.8 & 93.7 
    & \textit{71.6}$^*$ & \textit{70.1}$^*$ 
    & 48.9 & 44.6 
    & 94.5 & 94.5 \\
    $-$ Statistics compensation
    & 97.9 & 97.9 
    & \textit{88.8}$^*$ & \textit{85.3}$^*$
    & 98.2 & 98.2 
    & 93.6 & 93.4
    & 74.6 & 74.0 
    & 50.4 & 45.6 
    & \textit{89.5}$^*$ & \textit{89.5}$^*$ \\
    \bottomrule
  \end{tabular}
  }
  \label{tab:abalation result}
\end{table*}

\section{Validation of the STEP Structure}

\subsection{Baseline Comparison}
This section compares STEP against current SOTA time series structures in handling heterogenous scientific signals. Results are shown in Table~\ref{tab:baseline struc}. All models are trained from scratch to ensure fair comparison.

PatchTST demonstrates competitive performance on datasets characterized by relatively short sequences or strong local regularities (\textit{e.g.}, LEAVES and SleepEDF), confirming the effectiveness of patch-level representations for capturing localized temporal patterns. However, its performance degrades markedly on long-range or weakly periodic signals such as GWOSC and MarmAudio. This suggests that fixed patching schemes impose a rigid temporal granularity, which limits adaptability across heterogeneous scientific signals with vastly different temporal scales.

Informer exhibits consistently weaker performance across most tasks, with particularly poor results on GWOSC and MarmAudio. While its ProbSparse attention mechanism is effective for reducing computational cost in forecasting scenarios, it appears less suitable for discriminative scientific tasks where fine-grained temporal cues and global context must be jointly preserved. This highlights a structural mismatch between forecasting-oriented sparse attention and general-purpose representation learning.

Moirai shows strong performance on GWOSC, STEAD, and MarmAudio, and surpass TimeMoE on all downstream tasks. It is noted that although both designed for general time series forecasting, Moirai adopts an encoder-based structure\footnote{Moirai 1.1 used in this paper is encoder-based while its subsequent Moirai 2.0 uses decoder-only structure.} while TimeMoE uses a decoder-only structure.
This indicates that encoder-only architectures are better at effectively adapting their representations to scientific data. 

STEP achieves overall best performance across downstream tasks, ranking first on five of them and second on the rest. Effectiveness of the key design components are discussed in the following section.

\subsection{Ablation Studies}
We conduct ablation studies to evaluate the contribution of the two key components of the STEP encoder. Results are shown in Table~\ref{tab:abalation result}.

First, the learnable adaptive patching module is replaced by a fix length patching strategy. 
This variant employs rule-based patching, leaving sequences shorter than $T_\text{thres}$ unchanged and downsampling longer sequences to a fixed length of $T_\text{thres}$. The downsampling ratio is set as the sequence length divided by $T_\text{thres}$, rounded up:
$\lceil \frac{L}{T_\text{thres}} \rceil$,
where $L$ is the original sequence length. $T_\text{thres}$ is set to 200 here. A marked performance drop is observed on GWOSC and SleepEDF, which have long sequences and challenging tasks. For short sequences, adaptive patching learns very small patches, effectively resembling the rule-based approach.

We then remove the concatenation of per-sample mean and standard deviation statistics from the input representation. A distinct performance decrease is observed for LEAVES and RadSeg, where sequences are fairly short. In these cases, per-sample normalization alone is unreliable, as the computed mean and standard deviation carry little statistical significance. Appending the mean and standard deviation as auxiliary features mitigates this issue and preserves informative signal statistics.

\section{Cross Domain Knowledge Distillation}
\begin{figure}[t]
  \centering
  \includegraphics[width=0.9\columnwidth]{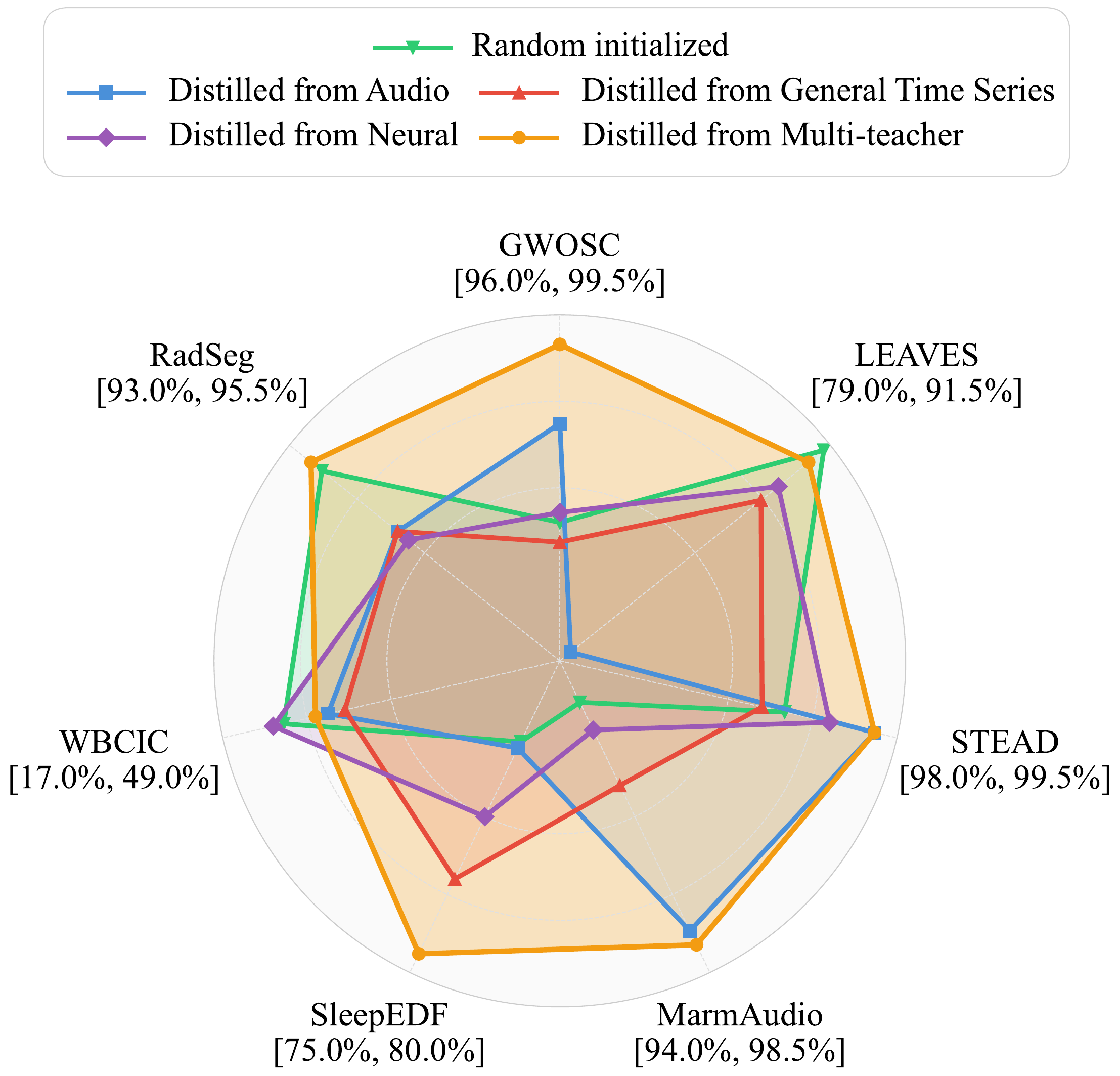}
    \caption{
      F1 scores on the downstream scientific time series tasks for the STEP encoder distilled from different teacher models.
    }
    \vspace{-1ex}
    \label{fig: radar}
\end{figure}

Based on the results in Section~\ref{sec: transfer}, we select SPEAR as the teacher from audio domain, TimeMoE as the teacher from general time series domain, and BrainOmni as the teacher from neural signal domain. Due to TimeMoE’s input length constraint, each sequence is truncated to 2048 time step segment. The target representation of BrainOmni is formed by concatenating its final compressed channel output.
A comparison between the STEP encoder trained from random initialization and that pretrained via distillation with different teacher models is presented in Figure~\ref{fig: radar}.

\textbf{Effectiveness of distillation on teacher-favoured tasks.}
We first observe that knowledge distillation consistently improves student performance on tasks where the teacher model itself exhibits strong downstream performance.
For instance, when distilled from SPEAR, the student encoder achieves  gains on GWOSC, STEAD, and MarmAudio, all of which are tasks where the SPEAR teacher demonstrates high accuracy. Similarly, distillation from TimeMoE  yields clear improvements on  SleepEDF. 
However, distillation from neural models yields counter-intuitive results. Although the neural teacher performs strong on WBCIC, the student encoder distilled from it shows limited gain, suggesting that this task may be particularly challenging or that neural foundation models are less effective as distillation teachers.

\textbf{Limited benefits on tasks where the teacher model struggles.}
In contrast, for tasks where the teacher model performs poorly, such as LEAVES, RadSeg, and WBCIC for SPEAR, the distilled encoder does not consistently outperform its random initialized counterpart. In these cases, performance is largely determined by the student’s own architectural inductive biases rather than the distilled knowledge.
An illustrative case is the MarmAudio dataset. While TimeMoE shows limited performance due to its fixed input length constraint during distillation, STEP encoder, free from such constraints, still benefits from TimeMoE supervision and achieves stronger downstream results after finetuning. This shows that distillation effectiveness is jointly determined by teacher quality and student capacity.

\textbf{Cross-teacher complementarity.}
Figure~\ref{fig: radar} reveals clear specialization among teacher models. SPEAR is particularly effective on long-range audio-like signals such as marmoset vocalization (MarmAudio), whereas TimeMoE excels on structured periodic and low-frequency time series such as sleeping EEG (SleepEDF). 
We observe that student gains generally reflect the relative strengths of individual teachers, which motivates leveraging multiple teachers to capture complementary knowledge.

\textbf{Multi-teacher distillation.} Multi-teacher distillation improves overall performance by integrating complementary strength from different teachers. As shown in Figure~\ref{fig: radar},
the multi-teacher setting achieves the most balanced performance across datasets, consistently matching or outperforming single-teacher distillation. While its gains over the best individual teacher are sometimes modest, multi-teacher distillation effectively avoids the degradation observed when a single teacher is poorly aligned with the target task. This indicates that aggregating heterogeneous teachers enables the student encoder to learn domain-agnostic representations that are more robust to teacher-specific biases. In addition, we observe that multi-teacher distillation does not necessarily outperform random initialization on tasks where none of the teachers performs well (\textit{e.g.}, LEAVES), indicating that scientific time series tasks differ fundamentally from well-established time series domains and that more tailored modelling approaches remain to be explored.

Overall, these results demonstrate that (i) distillation is most effective when the teacher model is well-aligned with the downstream task; (ii) student capacity plays an important role when the teacher struggles; (iii) leveraging multiple different teachers is beneficial for pretraining a scientific time series encoder; and (iv) scientific time series tasks are fundamentally different from existing time series domains, and more tailored methods remain to be explored.

\section{Conclusions}
In this work, we investigate the challenges of modelling scientific time series, which are characterized by extreme heterogeneity and data sparsity. We first analyze the transferability of foundation models pretrained on related time series domains such as audio, general time series, and neural signals and reveal their complementary strengths and limitations for scientific tasks. Based on these insights, we propose STEP encoder that combines adaptive patching, statistics compensation, and cross-domain distillation to integrate knowledge from multiple foundation models. Experiments on seven diverse scientific time series tasks demonstrate that STEP establishes an effective pretraining paradigm, consistently improving performance across domains. 
Our results highlight that scientific time series are fundamentally different from well-established time series domains while cross-domain foundation models hold great potential to accelerate representation learning for scientific time series, taking a STEP toward domain-adaptive pretraining and unified scientific signal modelling.

\section*{Impact Statements}
This paper presents work whose goal is to advance the field of machine learning. There are many potential societal consequences of our work, none of which we feel must be specifically highlighted here.

\bibliography{example_paper}
\bibliographystyle{icml2026}

\newpage
\appendix
\onecolumn

\end{document}